\documentclass[letterpaper]{article} 
\usepackage{aaai2027}  
\usepackage[hyphens]{url}  
\usepackage{graphicx} 
\usepackage{natbib}  
\usepackage{caption} 
\usepackage{algorithm}
\usepackage{algorithmic}
\usepackage{tikz}
\usepackage{graphicx, verbatim}
\usepackage{booktabs}
\usetikzlibrary{calc, positioning, matrix}
\usetikzlibrary{decorations.pathreplacing}
\usetikzlibrary{shapes.geometric, arrows.meta}
\usepackage{twemojis}
\usepackage{pifont}
\usepackage{amsmath}
\usepackage{amssymb}
\usepackage{dsfont}
\usepackage{makecell}

\usepackage{newfloat}
\usepackage{listings}
\DeclareCaptionStyle{ruled}{labelfont=normalfont,labelsep=colon,strut=off} 
\floatstyle{ruled}
\newfloat{listing}{tb}{lst}{}
\floatname{listing}{Listing}

\usepackage{booktabs}

\title{AAAI Press Anonymous Submission\\Instructions for Authors Using \LaTeX{}}
\author{
    Written by AAAI Press Staff\textsuperscript{\rm 1}\thanks{With help from the AAAI Publications Committee.}\\
    AAAI Style Contributions by Peter Patel Schneider,
    Sunil Issar,\\
    J. Scott Penberthy,
    George Ferguson,
    Hans Guesgen,
    Francisco Cruz\equalcontrib\corresponding,
    Marc Pujol-Gonzalez\equalcontrib\corresponding
}
\affiliations{
    \textsuperscript{\rm 1}Association for the Advancement of Artificial Intelligence\\

    1101 Pennsylvania Ave, NW Suite 300\\
    Washington, DC 20004 USA\\
    proceedings-questions@aaai.org
}

\title{Does Vision-Language Pretraining Granularity Matter? A Controlled Evaluation
of Vision-Language Objectives Across Chest X-Ray Interpretation Tasks}
\author {
    Denis Musinguzi\textsuperscript{\rm 1}\corresponding,
    Andrew Katumba\textsuperscript{\rm 2}\corresponding,
    Prasenjit Mitra\textsuperscript{\rm 1}
}
\affiliations {
    \textsuperscript{\rm 1}Carnegie Mellon University\\
    \textsuperscript{\rm 2}Makerere University\\
    dmusingu@andrew.cmu.edu, andrew.katumba@mak.ac.ug, pmitra@andrew.cmu.edu
}

\begin{document}

\maketitle

\begin{abstract}
Vision-language pretraining objectives differ in the spatial granularity of their supervision, yet the implications of this distribution for chest X-ray interpretation remain underexplored. We present a controlled study that isolates the pretraining objective: holding the encoder and pretraining data fixed, we train nine objectives spanning global and local contrastive learning, captioning, and their combinations, and evaluate across five chest X-ray tasks of increasing spatial granularity. We find that (i) pretraining granularity aligns with task granularity at the extremes, with local objectives leading on abnormality detection and global objectives on classification; (ii) local objectives are surprisingly competitive on global-level generation and question-answering tasks; (iii) the merits of captioning and contrastive learning reverse across granularity levels; and (iv) among combinations, mixing captioning and contrastive supervision is strongest on classification and in-distribution generation, while pairing two captioning objectives generalizes best on zero-shot report generation. These results show that no single objective is universally optimal, and that the interaction of objective type, granularity, and task governs downstream performance.

\end{abstract}

\begin{figure*}[t]
\begin{tikzpicture}
    \draw[->, very thick] (0,0) -- (17.5,0);

    \draw[decorate,decoration={brace, mirror, amplitude=5pt}]
    (0.2,-0.25) -- (3.0,-0.25)
    node[midway,below=6pt,align=center]
    {\textbf{Global-level}\\\footnotesize Image-level};

    \draw[decorate,decoration={brace, mirror, amplitude=5pt}]
    (3.2,-0.25) -- (8.9,-0.25)
    node[midway,below=6pt,align=center]
    {\textbf{Weak spatial}\\\footnotesize holistic};

    \draw[decorate,decoration={brace,mirror, amplitude=5pt}]
        (15.5,-0.25) -- (17.5,-0.25)
        node[midway,below=6pt,align=center]
        {\textbf{Region-level}\\\footnotesize Explicit localization};
    
    \draw[decorate,decoration={brace, mirror, amplitude=5pt}]
        (9.0,-0.25) -- (15.0,-0.25)
        node[midway,below=6pt,align=center]
        {\textbf{Relational}\\\footnotesize Temporal comparison};

    \node at (15.5,0.16) {Spatial Granularity};

    \node at (1.5,2.0) {
        \begin{tikzpicture}
            \node (class) {\includegraphics[width=2.5cm]{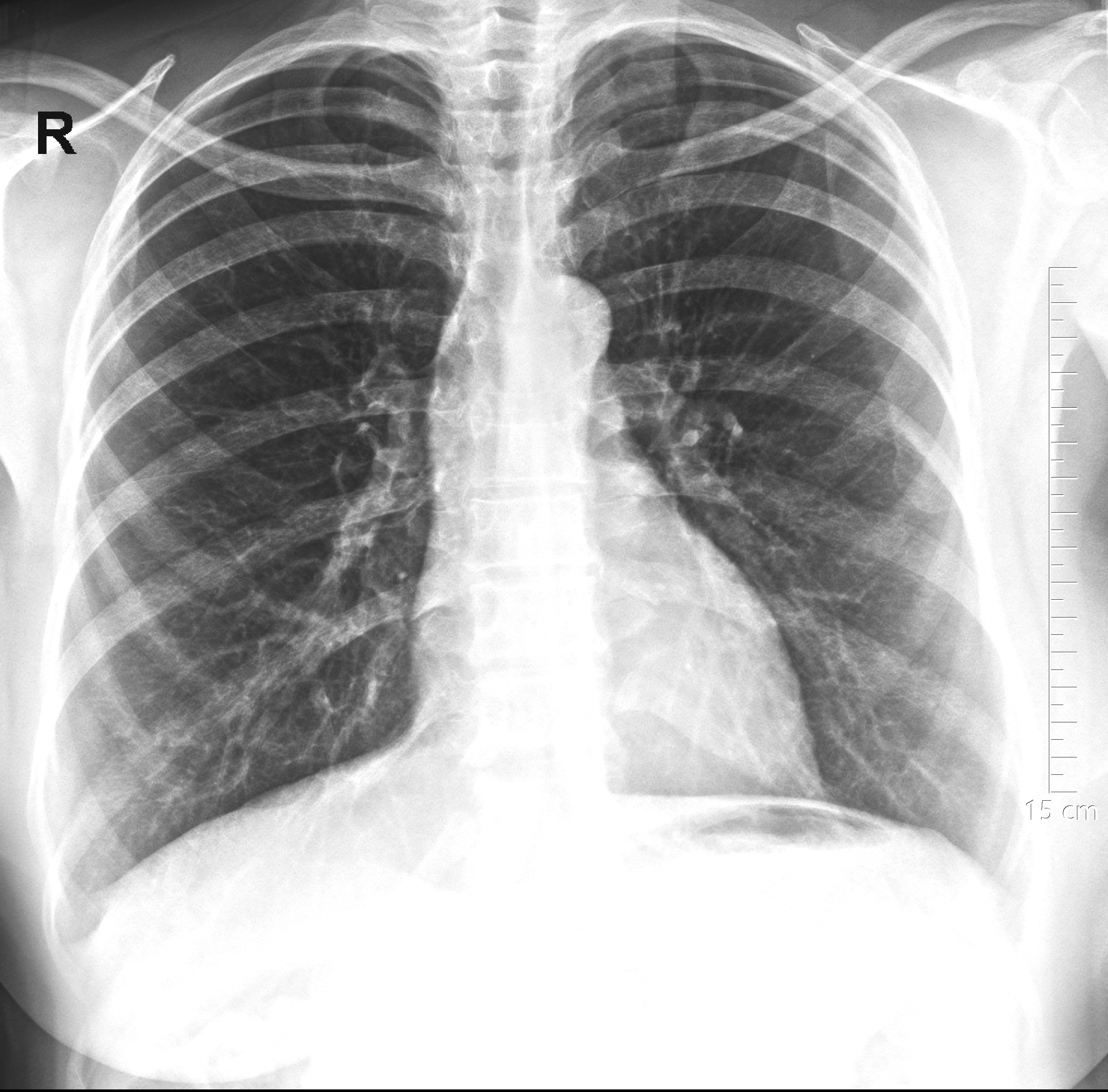}};
            \node[draw, rounded corners, fill=gray!10, text width=2.5cm, align=justify, font=\footnotesize, inner sep=3pt] at ([yshift=0.5cm]class.north)
            {\textbf{Classification}};
        \end{tikzpicture}
    };

    \node at (16.3,2.0) {
        \begin{tikzpicture}
            \node (abnorm) {\includegraphics[width=2.5cm]{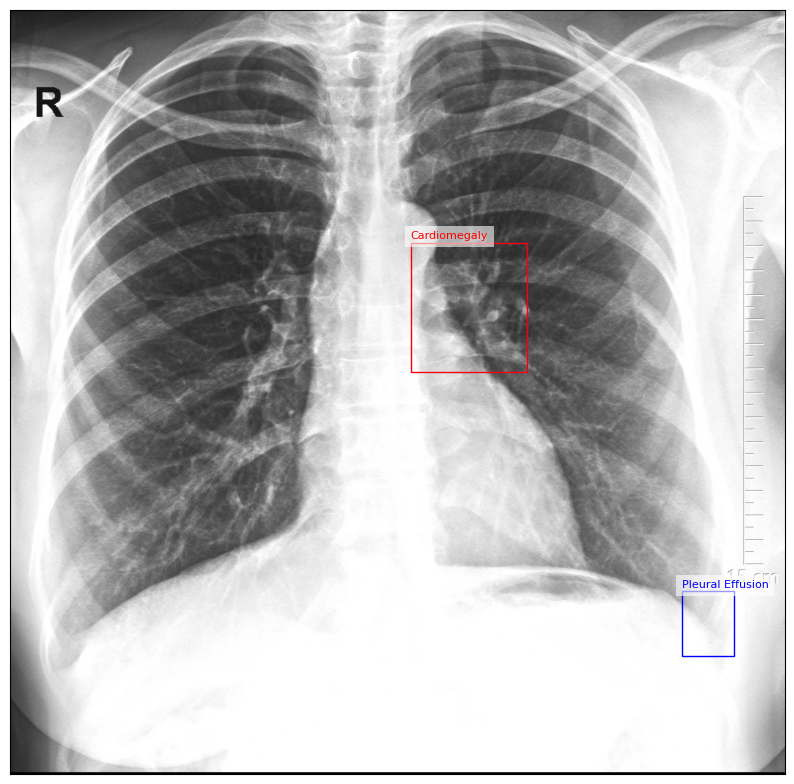}};
            \coordinate (adtop) at (abnorm.north);
            \node[draw, rounded corners, fill=gray!10, text width=2.5cm, align=justify, font=\footnotesize, inner sep=3pt] at ([yshift=0.5cm]adtop.north)
            {\textbf{Abnormality Detection}};
        \end{tikzpicture}
    };



    \node at (4.6,3.4) {
        \begin{tikzpicture}
            \node (rrg) {\includegraphics[width=2.5cm]{Figures/1.jpg}};
            
            \coordinate (imgtop) at (rrg.north);
            \node[draw, rounded corners, fill=gray!10,text width=3cm, align=justify, font=\footnotesize, inner sep=3pt]
            at ([yshift=1.95cm]imgtop)
        {\small \textbf{Report Generation}: The lungs are clear without focal consolidation, pleural effusion, or pneumothorax. No pulmonary edema is identified. The cardiomediastinal silhouette is within normal limits.};
        \end{tikzpicture}
    };

    \node at (12.15,2.8) {
    \begin{tikzpicture}
        \node (img1) {\includegraphics[width=2.5cm]{Figures/1.jpg}};
        \node (img2) [right=2mm of img1] {\includegraphics[width=2.5cm]{Figures/cardiomegaly.jpeg}};

        \coordinate (mid) at ($(img1.north)!0.5!(img2.north)$);

        \node[draw, rounded corners, fill=gray!10, text width=4.5cm, align=justify, inner sep=4pt, anchor=south]
        at ([yshift=0.2cm]mid)
        {\small \textbf{Medical Difference VQA}: What has changed compared to the reference image?  The main image has an additional finding of atelectasis than the reference image.};
    \end{tikzpicture}
    };

    \node at (7.7,2.5) {
        \begin{tikzpicture}
            \node (vqa) {\includegraphics[width=2.5cm]{Figures/cardiomegaly.jpeg}};
            \coordinate (vqatop) at (vqa.north);
            \node[draw, rounded corners, fill=gray!10, text width=2.3cm, align=justify, font=\footnotesize, inner sep=3pt]
            at ([yshift=0.75cm]vqatop)
            {\small \textbf{VQA:} Are there any abnormalities in the image? Cardiomegaly.};
        \end{tikzpicture}
    };
\end{tikzpicture}
\caption{Five downstream chest X-ray interpretation tasks with increasing spatial granularity requirements.}
\label{fig:spatial-granularity}
\end{figure*}

\section{Introduction}
Chest X-rays are the most widely used medical imaging procedure worldwide, valued for their accessibility and ease of use~\cite{rajpurkar2017chexnet}. However, growing demand has increased the workload on the limited number of radiologists who interpret these images. Radiologists perform a diverse range of tasks depending on the clinical context, from identifying abnormalities to localizing findings and generating detailed diagnostic reports~\cite{akhter2023ai}.
Machine learning models have shown considerable promise in chest X-ray interpretation, offering an opportunity to alleviate this burden. This is especially valuable in low-resource settings, where radiologist shortages are most acute.

Despite impressive benchmark results, clinical deployment of these models remains limited~\cite{dutta2025vision}.
Medical vision-language models typically employ frozen vision encoders pretrained with contrastive learning~\cite{radford2021learning}, applying the same representations across all chest X-ray interpretation tasks. A fundamental challenge is that these encoders often fail to extract sufficiently rich and task-relevant representations~\cite{tong2024eyes}. We argue that this is not merely incidental but a primary bottleneck on downstream performance, governed by the level of visual granularity encoded in the learned representations.

Chest X-ray interpretation tasks exhibit substantial variation in their granularity requirements, which can be arranged along a spectrum of spatial granularity as shown in Figure \ref{fig:spatial-granularity}. At one extreme, global-level tasks such as disease classification rely on high-level semantic understanding of the entire image. At the other end, local-level tasks such as abnormality detection and phrase grounding require spatially-grounded representations that focus on specific anatomical regions. Other tasks, such as report generation, visual question answering (VQA), and Medical Difference VQA~\cite{Hu_2023} fall between the two extremes, requiring both global and local representations in varying degrees.

Vision encoders are pretrained using vision-language objectives that vary along the same axis. Global objectives, such as image-text contrastive learning~\cite{radford2021learning} and image captioning~\cite{tschannen2023image}, incentivize learning coarse semantic representations aligned with global image content. While effective for high-level understanding tasks, these objectives provide limited supervision for learning spatially precise visual representations. In contrast, local objectives, such as referring expression comprehension, phrase grounding, and local contrastive learning, incorporate region-level supervision that explicitly aligns regional captions with local image regions, resulting in more spatially grounded representations~\cite{wan2024locca}. Despite this parallel, the extent to which pretraining granularity affects downstream performance across the task spectrum remains underexplored

In this work, we systematically investigate a fundamental question: \textbf{Do local vision-language pretraining objectives learn visual representations that are better suited for spatially grounded chest X-ray interpretation than global objectives, and does this advantage increase as tasks demand finer spatial granularity?} To answer this question, we perform a controlled empirical study in which we hold the vision encoder architecture fixed and vary only the pretraining objective. We evaluate the resulting representations across five downstream chest X-ray interpretation tasks: classification, report generation, VQA, medical difference VQA, and abnormality detection. 
By isolating the pretraining objective as the only experimental variable, we characterize how different vision-language objectives shape visual representations across increasingly demanding chest X-ray interpretation tasks. 

Our results support the hypothesis that localized pretraining benefits spatially-grounded tasks, with local objectives consistently outperforming global ones on abnormality detection, and global objectives edging out local ones on classification. However, the relationship between pretraining granularity and task granularity on tasks with intermediate granularity requirements is more nuanced. Local objectives are competitive on global-level tasks, outperforming global objectives on standard natural language metrics for report generation and visual question answering, while global captioning retains an advantage on clinically grounded RadGraph-F1 evaluation. We further find that the relative merits of captioning versus contrastive learning depend on the granularity level at which the objective is defined. Global captioning outperforms global contrastive learning, while local contrastive learning outperforms local captioning across most tasks. Among combined objectives, Local Contrastive and Global Captioning and Local Contrastive and Global Contrastive emerge as the strongest, with local contrastive learning proving to be a uniquely non-redundant signal whose presence consistently distinguishes the strongest combinations from the weakest.

\section{Related Work}
\subsection{Vision-Language Pretraining}
Traditionally, vision encoders were trained using supervised learning, which requires large amounts of manually annotated data and is therefore costly to scale. Inspired by the success of self-supervised learning in natural language processing, computer vision increasingly shifted toward leveraging large collections of image-text pairs available on the web. This shift motivated the development of vision-language pretraining methods that learn aligned representations of images and text.
Early approaches, such as CLIP~\cite{radford2021learning}, demonstrated that contrastive learning can produce vision encoders whose performance rivals that of supervised models. Subsequent methods, including SigLIP~\cite{zhai2023sigmoid}, further validated the effectiveness of image-text alignment objectives. 
Generative objectives based on image captioning~\cite{tschannen2023image} have since emerged as a strong alternative to contrastive pretraining, particularly for multimodal downstream tasks. More recent work~\cite{xiao2024florence, wan2024locca} has explored combining multiple pretraining objectives to capture information at different levels of semantic granularity within a shared vision encoder.

\subsection{Global and Localized Vision-Language Alignment}
Global alignment objectives~\cite{radford2021learning, zhai2023sigmoid, tschannen2023image} learn correspondences between entire images and their associated reports or captions. Methods such as contrastive learning~\cite{radford2021learning} and image captioning~\cite{tschannen2023image} operate at the image level, encouraging the model to learn semantic representations that capture the overall content of an image. While these approaches have demonstrated strong performance across a variety of downstream tasks, they do not explicitly model the relationship between specific image regions and textual descriptions.

Local alignment objectives~\cite{wan2024locca} instead induce correspondences between image regions and localized textual references, through tasks such as automatic referring expression comprehension, grounded captioning, and region-level contrastive learning. By providing explicit spatial supervision, these objectives encourage spatially grounded representations that may be better suited to localization and fine-grained reasoning. Recent work~\cite{huang2021gloria} has further shown that combining global and local objectives improves performance on multimodal tasks, suggesting that the spatial granularity of the pretraining signal plays an important role in the quality of the learned representations.

\section{Method} \label{sec:method}
Our goal is to investigate how the spatial granularity of vision-language pretraining objectives affects downstream chest X-ray interpretation. To isolate the effect of the objective, all models share the same vision encoder architecture, CLIP ViT-L/14\footnote{https://huggingface.co/openai/clip-vit-large-patch14}, and are pretrained on the same dataset, varying only in the alignment objective used.


\subsection{Global Alignment Objectives} 
\subsubsection{Contrastive Learning} \label{contrastive_learning_section}
Contrastive learning aligns images and text at the global level. Given a mini-batch $\mathcal{B}=\{(I_1, T_1), (I_2, T_2), ...\}$ of image-report pairs, the contrastive learning objective encourages embeddings of matching pairs $(I_i, T_i)$ to align with each other, while pushing embeddings of unmatched pairs $(I_i, T_{j\neq i})$ apart. 
A vision encoder ${f(\cdot)}$ and a text encoder $g(\cdot)$ are trained to minimize the following objective:
\begin{equation}
    - \frac{1}{2|\mathcal{B}|}\sum^{|\mathcal{B}|}_{i=1} \left(\text{log} \frac{e^{tx_i\cdot y_i}}{\sum^{|B|}_{j=1}e^{tx_i\cdot y_j}} + \text{log} \frac{e^{tx_i\cdot y_i}}{\sum^{|B|}_{j=1}e^{tx_i\cdot y_j}} \right)
\end{equation}
where $\text{x}_i = \frac{f(I_i)}{||f(I_i)||_2}$ and $y_i=\frac{g(T_i)}{||g(T_i)||_2}$.

We use a CLIP ViT-L/14 vision encoder for image representation and a bidirectional Transformer encoder for text. The text encoder consists of 6 layers with 8 attention heads, a hidden dimension of 512, a feed-forward dimension of 2048, and a dropout rate of 0.1. We use the GPT-2~\cite{radford2019language} tokenizer with a vocabulary size of 50,257 and a maximum sequence length of 128. We utilize learned positional embeddings, and a learned [CLS] token is prepended at position 0 to aggregate sequence-level information.
Both the vision and text encoders are trained from scratch. 
\subsubsection{Contrastive Learning with Sigmoid Loss}
The sigmoid-based loss processes every image-text pair independently, essentially casting the problem as a binary classification task on the dataset of all pair combinations. Matching pairs $(I_i, T_i)$ are assigned positive labels while all other pairs $(I_i, T_{j\neq i})$ are assigned negative labels. The loss is defined as follows:
\begin{equation}
    - \frac{1}{\mathcal{B}} \sum^{|B|}_{i=1} \sum^{|B|}_{j=1} \text{log} \frac{1}{1+e^{z_{ij}(-tx_i \cdot y_j + b)}}
\end{equation}
where $z_{ij}$ is the label for a given image and text input, which equals 1 if they match and -1 otherwise. An additional learnable bias $b$ is used to alleviate the initial imbalance between positive and negative pairs. The bias $b$ and the temperature $t$ are initialized to -10 and 10, respectively, following the original work.
We use an identical setup to the one used for contrastive learning.
\subsubsection{Image Captioning}
Image captioning formulates vision-language pretraining as a conditional text generation task. Given an image, $I$, the model generates the corresponding report ($W = {w_1, \ldots, w_T}$) auto-regressively, predicting each token conditioned on the image and all previously generated tokens. 
The image captioning objective is defined as:
\begin{equation}
\mathcal{L}_{\text{Cap}}
=
-
\sum_{t=1}^{T}
\log
P\left(
w_t
\mid
w_{<t},
I
\right).
\label{eqn:cap}
\end{equation}
where $I$ denotes the input image, $w_{<t}$ represents the sequence of previously generated tokens, and $T$ is the report length. The objective maximizes the likelihood of the ground-truth report conditioned on the image.

We utilize a standard Transformer decoder architecture for report generation. The sequence of visual embeddings produced by the ViT is provided to the decoder through cross-attention layers, enabling the decoder to attend to image regions while generating text. We use the same decoder architecture and hyperparameter configuration as those employed for the contrastive-learning objectives. The entire model is trained from scratch.
\subsection{Local Alignment Objectives}
\subsubsection{Local Captioning}
We adopt a set of regional captioning tasks to learn local correspondences between text and image regions. Regions are defined by anatomical structures in the chest, such as the mediastinum and left lung, and each is represented by a bounding box paired with a caption. From these annotations, we formulate three local vision-language tasks. Grounded Captioning (GCAP) generates a regional caption given an image and a bounding box. Automatic Referring Expression (AREF) predicts a bounding box given an image and a regional caption. Conditional Automatic Referring Expression (CAREF) generates both a caption and a bounding box given an image and the anatomical name of a region.
We train the model on all three tasks jointly in a multitask setup~\cite{beyer2023study}. Each task is formulated as an autoregressive sequence generation problem and optimized with the objective in Equation~\ref{eqn:cap}, with a task-specific prompt prepended to indicate the desired task and the loss computed over the generated tokens only. The decoder follows the same setup as global captioning.


\subsubsection{Local Contrastive} 
We extend contrastive learning from the image-report level to the region-caption level. Instead of aligning entire chest X-ray images with reports, we align anatomically defined image regions with their corresponding phrases: for each phrase, we contrast the ViT patch tokens within its bounding-box region against the phrase embedding. Regions that share the same phrase text and anatomical name are treated as positive pairs, while all remaining region-phrase combinations are treated as negatives. 
We optimize this objective with the same formulation, Transformer encoder, and training setup as global contrastive learning, differing only in the granularity of the aligned pairs.

\subsection{Combined Local and Global Objectives}

\subsubsection{Local Contrastive Learning and Global Captioning}
A shared vision encoder is trained jointly on both objectives, with separate task-specific heads: a bidirectional transformer encoder for regional contrastive learning and a Transformer decoder for global captioning. To balance the two objectives, training is interleaved so that each is optimized on alternating steps, giving both an equal number of updates.

\subsubsection{Local Captioning and Global Captioning}
To incorporate global supervision with local captioning, we introduce captioning (CAP), which prompts the model to generate the full radiology report from the input image. All four tasks are trained jointly within a multitask autoregressive framework using a shared vision encoder and text decoder, with all four tasks sampled in equal proportion so that no single task dominates.

\subsubsection{Local Contrastive and Global Contrastive}
A shared vision encoder and text encoder are trained jointly on both objectives. To balance the objectives, training is interleaved such that each objective is optimized at alternating steps, giving both an equal number of updates.

\subsubsection{Local Captioning and Global Contrastive}
A shared vision encoder is trained jointly on both local captioning and global contrastive objectives, with separate task-specific heads: a bidirectional encoder for global contrastive learning and a transformer decoder for local captioning. Training of the two objectives is interleaved to balance the contribution from the objectives.

\section{Experiment}\label{sec:experiments}
\subsection{Task Setup}
\label{sec:task}
We use MIMIC-CXR~\cite{johnson2019mimic} for vision-language pretraining, retaining only PA and AP views, excluding studies missing a findings or impression section, and removing references to prior imaging. After filtering, 242,575 studies remain across the official splits. We partition the training set into two disjoint subsets: 100,000 studies for pretraining and 99,341 for downstream training of report generation, VQA, and medical difference VQA. Images are resized to $384\times384$ and converted to RGB. In all downstream experiments, we freeze the encoder and train only a task-specific head, isolating the effect of the pretraining objective.

For objectives requiring local supervision, we use Chest ImaGenome~\cite{wu2021chest}, which augments MIMIC-CXR with region-level bounding boxes and report phrases. Since identical phrases often map to multiple regions, we assign each phrase to a single anatomical structure, choosing the most specific region when nested regions apply; negative findings are assigned to the region where the pathology would typically manifest. Bounding boxes are represented as $[x_{\min}, y_{\min}, x_{\max}, y_{\max}]$, normalized by image dimensions and scaled by 1000.

All models use the CLIP ViT-L/14~\cite{radford2021learning} vision encoder, trained from scratch alongside a text module instantiated as a bidirectional Transformer encoder for contrastive objectives or an autoregressive decoder for captioning objectives. Both text module variants use 6 layers, 8 attention heads, hidden dimension 512, feed-forward dimension 2048, and dropout 0.1, with the GPT-2 tokenizer (50,257 tokens), learned positional embeddings, and a learned \texttt{[CLS]} token for sequence-level representations. Models are trained for 50 epochs with AdamW, a learning rate of $1\times10^{-4}$, and cosine annealing; we select the checkpoint with the lowest validation loss.

The resulting encoders are evaluated on six downstream tasks spanning the granularity spectrum introduced in the Introduction Section, each described in the following subsections. The performance for all generative tasks is measured using BLEU-1/4, ROUGE-1/L, RadGraph-F1, and BERTScore. The performance for classification is measured using AUC while that of abnormality detection is measured using mAP@0.4 and mAP@0.5. We report the mean and standard deviation over three seeds. We perform paired bootstrap resampling~\cite{berg2012empirical} with Benjaminin-Hochberg FDR correction~\cite{benjamini1995controlling} across all pairwise comparisons; full results are presented in the technical supplement.

\subsubsection{Multilabel Classification}
Our primary evaluation protocol for classification is linear probing. In this setting, a single linear layer is trained on top of visual representations extracted from the vision encoder using binary cross-entropy (BCE) loss with SGD optimizer, a learning rate of 0.1, cosine annealing scheduler, a batch size of 512, and for 30 epochs. 
We utilize the NIH CXR-14 dataset~\cite{wang2017chestxray}, a 14-class multi-label dataset comprising 94,596 chest X-ray images from 30,805 patients. 
We strictly follow the official train and test splits.


\subsubsection{Abnormality Detection}

We train a DETR~\cite{carion2020end} detection head on the frozen features using a combination of classification, L1 bounding-box regression, and generalized IoU losses:
\begin{equation}
L = \mathrm{CE}(\text{bg\_weight}=0.1) + \lambda_1 L_1 + \lambda_g \mathrm{GIoU}
\end{equation}
where $\lambda_1 = 5$ and $\lambda_g = 2$. The head comprises a 2-layer Transformer encoder and a 3-layer decoder, each with 8 attention heads, hidden dimension 256, dropout 0.1, and pre-normalization throughout. It uses 20 learned object queries, a linear classification head over 15 classes (14 pathologies plus background), and a 3-layer MLP regression head.
We evaluate on VinDr-CXR~\cite{nguyen2022vindr}, an 18,000-image dataset with 14 pathology classes, split into 15,000 training and 3,000 test images. We train with AdamW for 50 epochs, using a learning rate of $1\times10^{-4}$, batch size 8, and cosine annealing. 





\subsubsection{Report Generation}
We train a text decoder on top of extracted vision features to generate a free-text radiology report. The decoder is trained on features from the 99,341 MIMIC-CXR training images excluded from the pretraining stage. The decoder architecture is described under task setup. It is trained from scratch using cross-entropy loss, following the objective in Eq.~\ref{eqn:cap}, with AdamW optimizer, a learning rate of $1\times10^{-4}$, a cosine annealing scheduler, a batch size of 64, and for 20 epochs. We evaluate performance on the MIMIC-CXR test set, the IU X-ray dataset, and the CheXpert Plus dataset, retaining only frontal chest radiographs to ensure consistency with MIMIC-CXR. Importantly, the decoder is not trained on IU X-ray or CheXpert Plus, so these constitute zero-shot evaluations. 

\subsubsection{Visual Question Answering}
We train a text decoder on the frozen vision features to answer questions about chest X-ray images, using the single-image subset of the Medical Difference VQA dataset~\cite{hu2023medicaldiff}. The questions span six categories — location, presence, abnormality, type, severity, and view — with 202,982 training and 25,689 test questions. To avoid data leakage, we retain only questions whose images were excluded from pretraining. The decoder architecture is described under Task Setup: at each layer, masked self-attention models the question and previously generated answer tokens, while cross-attention integrates the visual features. It is trained from scratch with cross-entropy loss, using AdamW for 10 epochs with a learning rate of $1\times10^{-4}$, batch size 64, and cosine annealing with a 5\% linear warm-up. 

\subsubsection{Medical Difference VQA}
We train a text decoder on the frozen vision features to describe longitudinal changes between a pair of chest X-rays acquired from the same patient at different time points. We use the ``difference'' subset of the Medical Difference VQA dataset, comprising 104,066 training, 12,934 validation, and 12,918 test questions. To prevent data leakage, we ensure that neither image in any test pair appears in the pretraining dataset. The frozen features from the two images are concatenated into a single cross-attention memory, with learned frame embeddings added to indicate whether each token originates from the reference or current image. In each decoder layer, masked self-attention models the question and previously generated answer tokens, while cross-attention integrates information from both images. The decoder architecture is described under Task Setup. It is trained from scratch with cross-entropy loss, using AdamW for 20 epochs with a learning rate of $1\times10^{-4}$, weight decay 0.01, batch size 20, and cosine annealing.



\begin{figure}[t]
    \centering
    \includegraphics[width=1.0\linewidth]{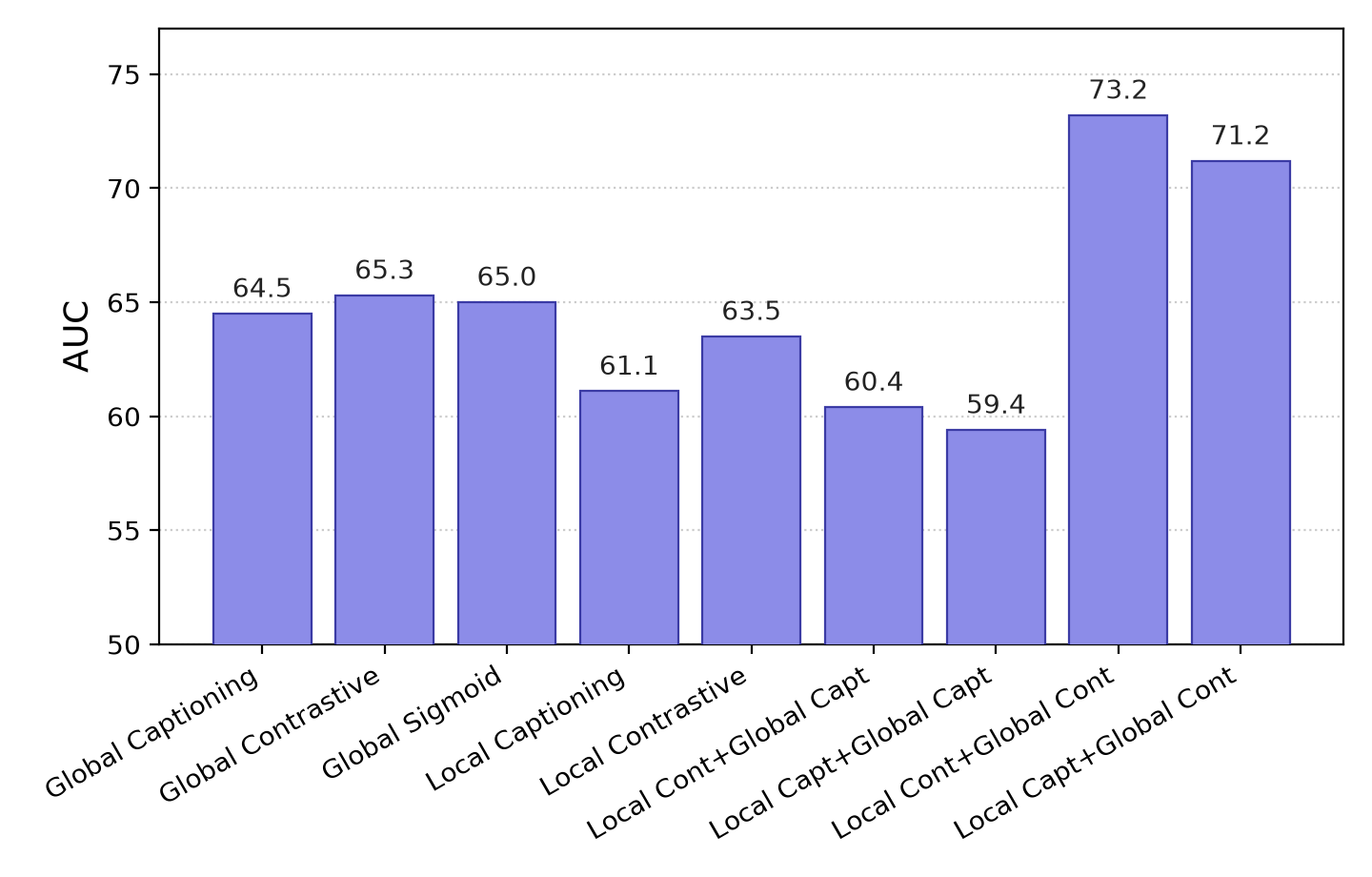}
    \caption{AUC scores on NIH CXR-14 classification dataset. Cont denotes Contrastive and Capt denotes Captioning.}
    \label{fig:nih_cxr14}
\end{figure}

\begin{figure}[ht]
    \centering
    \includegraphics[width=1.0\linewidth]{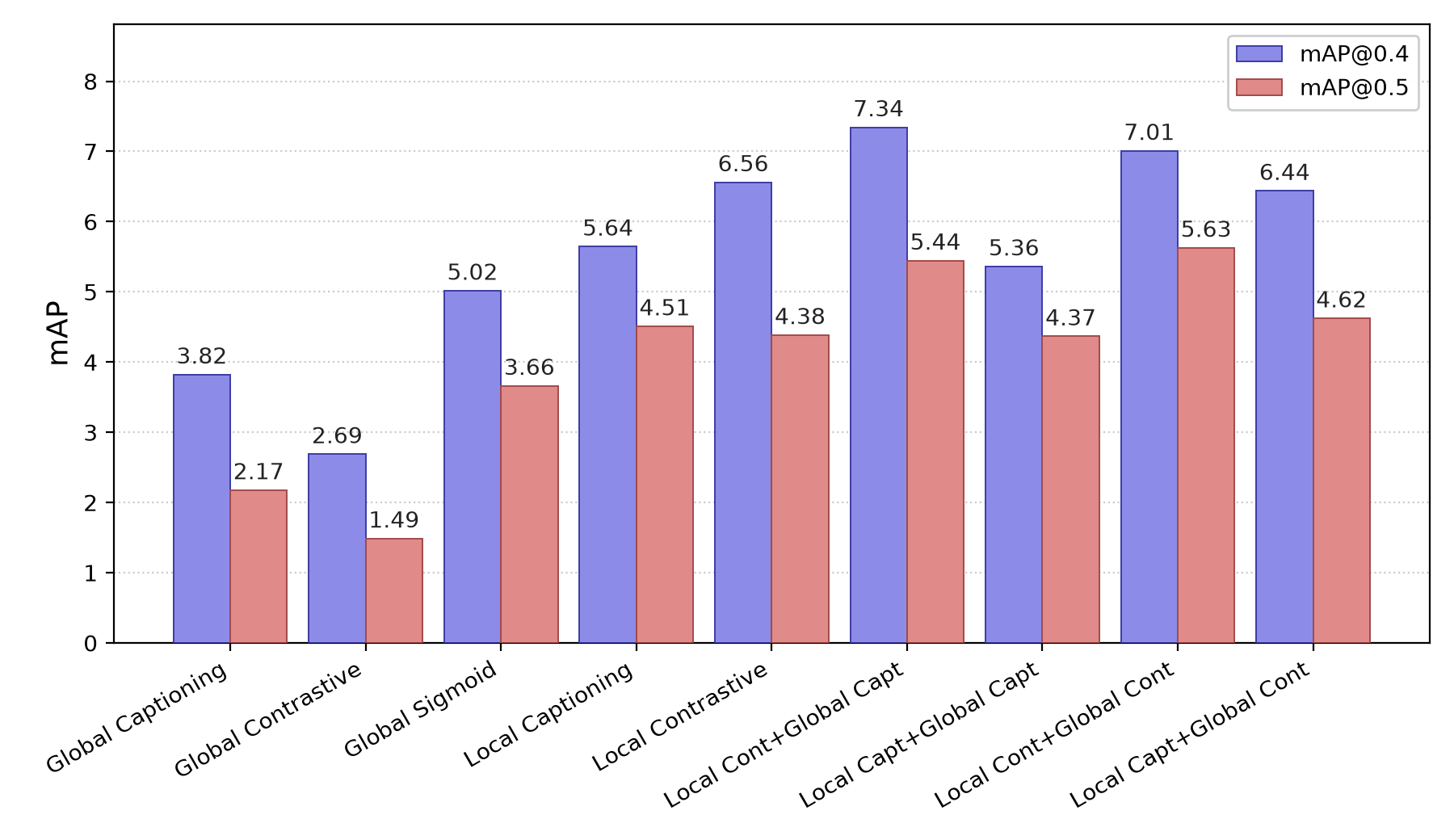}
    \caption{Evaluation results on the VinDr-CXR abnormality detection benchmark.}
    \label{fig:vindr-cxr-results}
\end{figure}

\begin{table*}[t]
\centering
\caption{Report generation performance across MIMIC-CXR, IU X-ray, and CheXpert Plus. IU X-ray and CheXpert Plus are zero-shot (the decoder is trained only on MIMIC-CXR). All results are percentages; best in \textbf{bold}, second-best \underline{underlined}.}
\label{tab:report_gen}
\setlength{\tabcolsep}{6pt}
\begin{tabular}{l|cccc}
\hline
Encoder & BLEU-4 & ROUGE-L & RadGraph & BERTScore \\
\hline
\multicolumn{5}{c}{\textit{MIMIC-CXR}} \\
\hline
Global Captioning & 3.08$\pm$0.03 & 15.89$\pm$0.05 & 16.24$\pm$0.09 & 84.68$\pm$0.04 \\
Global Contrastive & 2.92$\pm$0.03 & 15.43$\pm$0.05 & 15.35$\pm$0.09 & 84.58$\pm$0.04 \\
Global Sigmoid & 2.84$\pm$0.03 & 15.50$\pm$0.05 & 15.22$\pm$0.08 & 84.62$\pm$0.03 \\
Local Captioning & 3.12$\pm$0.05 & 16.10$\pm$0.05 & {16.34}$\pm$0.09 & 84.73$\pm$0.04 \\
Local Contrastive & 3.16$\pm$0.03 & {16.17}$\pm$0.05 & 16.24$\pm$0.08 & \underline{84.86}$\pm$0.03 \\
Local Contrastive + Global Captioning & \underline{3.24}$\pm$0.03 & \textbf{16.52}$\pm$0.03 & \textbf{16.76}$\pm$0.08 & \textbf{84.99}$\pm$0.03 \\
Local Captioning + Global Captioning & {3.19}$\pm$0.03 & 16.35$\pm$0.05 & 16.48$\pm$0.09 & 84.69$\pm$0.04 \\
Local Contrastive + Global Contrastive & 3.07$\pm$0.03 & 15.84$\pm$0.05 & \underline{16.69}$\pm$0.08 & 84.61$\pm$0.03 \\
Local Captioning + Global Contrastive &\textbf{3.26}$\pm$0.03&\underline{16.51}$\pm$0.05&16.59$\pm$0.09&84.81$\pm$0.03\\
\hline
\multicolumn{5}{c}{\textit{IU X-ray (zero-shot)}} \\
\hline
Global Captioning & 3.06$\pm$0.16 & 16.03$\pm$0.23 & \textbf{32.79}$\pm$0.46 & 84.90$\pm$0.06 \\
Global Contrastive & 3.12$\pm$0.16 & 15.96$\pm$0.23 & 30.32$\pm$0.42 & 85.43$\pm$0.06 \\
Global Sigmoid & 2.98$\pm$0.16 & 15.75$\pm$0.24 & \underline{31.93}$\pm$0.44 & 85.08$\pm$0.06 \\
Local Captioning & {3.21}$\pm$0.18 & 16.39$\pm$0.26 & 29.90$\pm$0.39 & \textbf{85.69}$\pm$0.06 \\
Local Contrastive & 3.15$\pm$0.16 & {16.48}$\pm$0.25 & 30.70$\pm$0.41 & \underline{85.47}$\pm$0.06 \\
Local Contrastive + Global Captioning & 1.98$\pm$0.15 & 15.33$\pm$0.24 & 29.59$\pm$0.40 & \textbf{85.69}$\pm$0.06 \\
Local Captioning + Global Captioning & \textbf{3.27}$\pm$0.16 & \textbf{16.77}$\pm$0.25 & 31.07$\pm$0.43 & 85.17$\pm$0.06 \\
Local Contrastive + Global Contrastive & 3.09$\pm$0.21 & 15.77$\pm$0.24 & 30.52$\pm$0.41 & 85.46$\pm$0.06 \\
Local Captioning + Global Contrastive &\underline{3.21}$\pm$0.16&\underline{16.62}$\pm$0.25&29.38$\pm$0.40&85.02$\pm$0.06
\\
\hline
\multicolumn{5}{c}{\textit{CheXpert Plus (zero-shot)}} \\
\hline
Global Captioning & 2.19$\pm$0.22 & 13.65$\pm$0.45 & \underline{18.48}$\pm$0.89 & 83.71$\pm$0.12 \\
Global Contrastive & 2.19$\pm$0.22 & 12.97$\pm$0.44 & 16.62$\pm$0.87 & 83.55$\pm$0.12 \\
Global Sigmoid & 2.06$\pm$0.22 & 13.26$\pm$0.44 & 16.01$\pm$0.83 & 83.66$\pm$0.11 \\
Local Captioning & \underline{2.77}$\pm$0.33 & 13.61$\pm$0.49 & 18.03$\pm$0.89 & 83.81$\pm$0.12 \\
Local Contrastive & 2.19$\pm$0.25 & 14.03$\pm$0.44 & 17.82$\pm$0.88 & \underline{83.97}$\pm$0.11 \\
Local Contrastive + Global Captioning & 1.98$\pm$0.21 & 13.73$\pm$0.43 & 18.15$\pm$0.92 & 83.88$\pm$0.12 \\
Local Captioning + Global Captioning & \textbf{3.01}$\pm$0.32 & \underline{14.72}$\pm$0.51 & \textbf{19.10}$\pm$0.86 & 83.81$\pm$0.13 \\
Local Contrastive + Global Contrastive & 2.20$\pm$0.22 & 14.04$\pm$0.44 & 17.85$\pm$0.88 & \textbf{83.98}$\pm$0.11 \\
Local Captioning + Global Contrastive &1.46$\pm$0.21&10.91$\pm$0.43&12.33$\pm$0.86&83.02$\pm$0.11\\
\hline
\end{tabular}
\end{table*}
\section{Results}

We evaluate nine pretraining objectives across five chest X-ray interpretation tasks spanning two granularity levels. Rather than a single objective emerging as a universal winner, the results reveal a nuanced interplay between objective type, granularity level, and task characteristics. 

\begin{table*}[tb]
    \centering
    \begin{tabular}{l|cccc}
    \hline
         Encoders & BLEU-4 & ROUGE-L & RadGraph & BERTScore \\
         \hline
         Global Captioning &
         24.03$\pm$1.00 & 70.12$\pm$0.26 &
         19.43$\pm$0.23 & {95.64}$\pm$0.04 \\
         
         Global Contrastive &
         20.88$\pm$0.73 & 64.72$\pm$0.27 &
         20.86$\pm$0.22 & 95.42$\pm$0.04 \\
         
         Global Sigmoid &
         21.57$\pm$0.73 & 59.64$\pm$0.29 &
         \underline{21.65}$\pm$0.23 & 94.89$\pm$0.05 \\
         \hline
         
         Local Captioning &
         26.39$\pm$0.86 & 70.08$\pm$0.27 &
         19.85$\pm$0.23 & 95.53$\pm$0.05 \\
         
         Local Contrastive &
         \textbf{28.55}$\pm$0.96 & {70.92$\pm$0.26} &
         20.58$\pm$0.23 & \underline{95.75}$\pm$0.04 \\
         \hline

         \makecell[l]{Local Contrastive + Global Captioning} &
         \underline{28.53}$\pm$0.89 & \textbf{71.92}$\pm$0.26 &
         20.62$\pm$0.23 & 95.62$\pm$0.05 \\
         
         \makecell[l]{Local Captioning + Global Captioning} &
         25.79$\pm$0.99 & 68.94$\pm$0.27 &
         20.70$\pm$0.23 & 95.63$\pm$0.05 \\
         
         \makecell[l]{Local Contrastive + Global Contrastive} &
         25.49$\pm$0.98 & 70.28$\pm$0.26 &
         21.09$\pm$0.23 & 95.38$\pm$0.05 \\
         Local Captioning + Global Contrastive &24.05$\pm$0.77&\underline{71.79}$\pm$0.26&\textbf{22.19}$\pm$0.24&\textbf{97.10}$\pm$0.04\\
         \hline
    \end{tabular}
    \caption{Single-image VQA results of vision encoders trained with different pretraining objectives evaluated on the Medical Difference VQA test set. All results are expressed as percentages. Best results are shown in \textbf{bold}, and second-best results are \underline{underlined}.}
    \label{tab:single_vqa}
\end{table*}

\begin{table*}[tb]
    \centering
    \begin{tabular}{l|cccc}
    \hline
         Encoder & BLEU-4 & ROUGE-L & RadGraph & BERTScore \\
         \hline
         Global Captioning &
         46.83$\pm$0.40 & 64.30$\pm$0.41 & 29.80$\pm$0.48 & 93.60$\pm$0.06 \\
         
         Global Contrastive &
         44.57$\pm$0.45 & 63.80$\pm$0.40 & 29.04$\pm$0.46 & 93.48$\pm$0.06 \\
         
         Global Sigmoid &
         44.90$\pm$0.40 & 63.25$\pm$0.40 & 29.22$\pm$0.47 & 93.53$\pm$0.06 \\
         \hline
         
         Local Captioning &
         43.68$\pm$0.38 & 63.93$\pm$0.41 & \textbf{30.31}$\pm$0.45 & 93.53$\pm$0.06 \\
         
         Local Contrastive &
         46.54$\pm$0.39 & 65.30$\pm$0.40 & 29.77$\pm$0.47 & \underline{93.74}$\pm$0.06 \\
         \hline
         
         \makecell[l]{Local Contrastive + Global Captioning} &
         \textbf{47.96$\pm$0.40} & \textbf{66.04}$\pm$0.41 & 28.98$\pm$0.48 & \textbf{93.79}$\pm$0.06 \\
         
         \makecell[l]{Local Captioning + Global Captioning} &
         44.91$\pm$0.40 & 64.02$\pm$0.40 & 26.77$\pm$0.45 & 93.49$\pm$0.06 \\
         
         \makecell[l]{Local Contrastive + Global Contrastive} &
         {47.06$\pm$0.40} & {65.79$\pm$0.41} & 
         \underline{29.96$\pm$0.48} & \textbf{93.79}$\pm$0.06 \\
         Local Captioning + Global Contrastive &\underline{47.37}$\pm$0.40&\underline{66.01}$\pm$0.42&29.61$\pm$0.47&93.72$\pm$0.06\\
         \hline
    \end{tabular}
    \caption{Evaluation results of vision encoders trained with different pretraining objectives on the difference subset of the Medical Difference VQA benchmark, where each question involves reasoning across a pair of images. All results are expressed as percentages. Best results are shown in \textbf{bold}, and second-best results are \underline{underlined}.}
    \label{tab:diff_vqa}
\end{table*}

\subsection{Pretraining granularity aligns with task granularity at the extremes}
The clearest support for the granularity hypothesis comes from the two tasks at opposite ends of the spatial granularity spectrum. On abnormality detection (VinDr-CXR, Figure~\ref{fig:vindr-cxr-results}), the most spatially demanding task, localized objectives outperform global ones: Local Contrastive (6.56 mAP@0.4) and Local Captioning (5.64) surpass all global single objectives, the best being Global Sigmoid (5.02). Conversely, on classification (NIH-CXR-14, Figure~\ref{fig:nih_cxr14}), a global task, global objectives win: Global Contrastive leads among single objectives (65.3 AUC), with Global Sigmoid (65.0) and Global Captioning (64.5) also surpassing the best local objective, Local Contrastive (63.5). This alignment, with local objectives winning on the localized task and global objectives on the global one, is consistent with the hypothesis that pretraining granularity should match task granularity. Two nuances stand out. On detection, the local advantage is strongest for Local Contrastive, while Local Captioning's margin over Global Sigmoid is modest. Within the global objectives on detection, Global Sigmoid substantially outperforms Global Contrastive (2.69) and Global Captioning (3.82), suggesting its pairwise binary loss provides a finer-grained signal than InfoNCE even at the image level. Detection and classification thus offer complementary evidence for the hypothesis at the two extremes; the intermediate tasks, examined next, reveal a more complex picture.

\subsection{Local objectives are competitive on global tasks}
Across report generation and VQA, local objectives consistently match or outperform global ones on NLG metrics. On MIMIC-CXR (Table~\ref{tab:report_gen}), Local Contrastive achieves the best BLEU-4 (3.16) and ROUGE-L (16.17) among single objectives, outperforming Global Captioning (3.08 and 15.89), with both differences statistically significant under paired bootstrap resampling. The pattern holds on IU X-ray (Table~\ref{tab:report_gen}), where Local Captioning leads on BLEU-4 (3.21) and Local Contrastive on ROUGE-L (16.48), each surpassing the best global objective. On CheXpert Plus (Table~\ref{tab:report_gen}), Local Captioning leads on BLEU-4 (2.77) and Local Contrastive on ROUGE-L (14.03). The local advantage is most pronounced on single-image VQA (Table~\ref{tab:single_vqa}), where Local Contrastive leads Global Sigmoid by nearly 7 BLEU-4 points (28.55 vs.\ 21.57), the largest margin between any two single objectives in the study, and is statistically significant. This advantage does not extend uniformly to RadGraph, however: differences there are generally small and inconsistent, with Global Captioning competitive on report generation but not consistently ahead on VQA. This suggests that the benefits of local pretraining are more clearly reflected in surface-level text quality than in clinical entity accuracy.

Two complementary explanations account for this competitiveness. First, local objectives may produce representations that subsume global semantics: by learning to attend to specific regions and findings, the encoder develops rich patch-level features that, when aggregated, yield stronger image-level representations than objectives optimized solely at the global level. Second, report generation and VQA, though nominally global, require some local understanding to produce clinically accurate outputs, since reports describe specific findings at specific anatomical locations and questions may probe specific regions. Local pretraining may therefore confer an implicit advantage even on global tasks by equipping the encoder with the fine-grained detail needed for accurate clinical description.

\subsection{Captioning and contrastive learning excel at different granularity levels}
Within the global category, captioning outperforms contrastive on most tasks, and the MIMIC-CXR significance results confirm that Global Captioning is significantly better than both Global Contrastive and Global Sigmoid across all metrics. Within the local category, the pattern reverses: Local Contrastive outperforms Local Captioning on most tasks, including detection (Figure~\ref{fig:vindr-cxr-results}),classification (Figure~\ref{fig:nih_cxr14}), VQA (Table~\ref{tab:single_vqa}), and difference VQA (Table~\ref{tab:diff_vqa}), and is competitive or leading on all three report generation datasets (Tables~\ref{tab:report_gen}). This interaction suggests that the relative merits of captioning and contrastive learning cannot be assessed independently of the spatial scale at which they are applied.

\subsection{Mixing captioning and contrastive objectives yields the strongest combinations}
On classification and in-distribution generation, combinations mixing captioning and contrastive are strongest. Local contrastive and global captioning leads on NLG metrics for MIMIC-CXR report generation (Table~\ref{tab:report_gen}), VQA (Table~\ref{tab:single_vqa}), and medical difference VQA (Table~\ref{tab:diff_vqa}), and its mirror image, Local captioning and global contrastive, is similarly strong, indicating that the benefit steps from mixing supervision types rather than the granularity assignment. On classification, the combinations with global contrastive learning lead with a considerable margin over single objective encoders underscoring the contrastive component's importance for discriminative tasks. Local captioning and global captioning departs from this pattern: it is worst on classification and on MIMIC-CXR RadGraph, yet strongest among combinations on the zero-shot IU X-ray and CheXpert Plus report generation datasets, revealing that pure captioning supervision generalizes best out-of-distribution even when it under performs in-distribution.

\subsection{Limitations}
Our findings are scoped to the frozen-encoder transfer setting and may not hold under full finetuning, which could alter the relative standing of the objectives. We evaluate a single architecture (CLIP ViT-L/14) at one pretraining scale, and most intermediate tasks derive from MIMIC-CXR, limiting out-of-distribution evidence. Finally, the local objectives require dense region-level annotations (bounding boxes paired with phrases), which are costly to obtain and unavailable for most datasets and imaging modalities, constraining the settings in which localized pretraining is practical.

\section{Conclusion}
We investigated how pretraining objectives affect five chest X-ray interpretation tasks spanning two levels of granularity. Our results partially support the hypothesis that granularity-matched pretraining helps: localized objectives lead on abnormality detection and global objectives on classification. The picture is more complex on intermediate tasks, where local objectives are competitive on report generation and VQA, outperforming global objectives on NLG metrics while global captioning retains an edge on RadGraph-F1. The merits of captioning versus contrastive learning also depend on granularity, with global captioning beating global contrastive but local contrastive beating local captioning across most tasks. Among combinations, mixing captioning and contrastive supervision is strongest on classification and in-distribution generation, while two captioning objectives generalize best on zero-shot report generation. These findings challenge the assumption that pretraining granularity maps cleanly onto task requirements, pointing instead to the interaction of objective type, granularity, and task as the primary determinant of downstream performance.

\bibliography{aaai2027}

@article{nguyen2022vindr,
  title={VinDr-CXR: An open dataset of chest X-rays with radiologist’s annotations},
  author={Nguyen, Ha Q and Lam, Khanh and Le, Linh T and Pham, Hieu H and Tran, Dat Q and Nguyen, Dung B and Le, Dung D and Pham, Chi M and Tong, Hang TT and Dinh, Diep H and others},
  journal={Scientific Data},
  volume={9},
  number={1},
  pages={429},
  year={2022},
  publisher={Nature Publishing Group UK London}
}

@article{johnson2019mimic,
  title={MIMIC-CXR, a de-identified publicly available database of chest radiographs with free-text reports},
  author={Johnson, Alistair EW and Pollard, Tom J and Berkowitz, Seth J and Greenbaum, Nathaniel R and Lungren, Matthew P and Deng, Chih-ying and Mark, Roger G and Horng, Steven},
  journal={Scientific data},
  volume={6},
  number={1},
  pages={317},
  year={2019},
  publisher={Nature Publishing Group UK London}
}

@article{hu2023medicaldiff,
  title={Medicaldiff-vqa: a large-scale medical dataset for difference visual question answering on chest x-ray images},
  author={Hu, Xinyue and Gu, Lin and An, Qiyuan and Zhang, Mengliang and Liu, Liangchen and Kobayashi, Kazuma and Harada, Tatsuya and Summers, R and Zhu, Yingying},
  journal={PhysioNet},
  volume={12},
  pages={13},
  year={2023}
}

@inproceedings{berg2012empirical,
  title={An empirical investigation of statistical significance in NLP},
  author={Berg-Kirkpatrick, Taylor and Burkett, David and Klein, Dan},
  booktitle={Proceedings of the 2012 joint conference on empirical methods in natural language processing and computational natural language learning},
  pages={995--1005},
  year={2012}
}

@article{benjamini1995controlling,
  title={Controlling the false discovery rate: a practical and powerful approach to multiple testing},
  author={Benjamini, Yoav and Hochberg, Yosef},
  journal={Journal of the Royal statistical society: series B (Methodological)},
  volume={57},
  number={1},
  pages={289--300},
  year={1995},
  publisher={Wiley Online Library}
}

@inproceedings{radford2021learning,
  title={Learning transferable visual models from natural language supervision},
  author={Radford, Alec and Kim, Jong Wook and Hallacy, Chris and Ramesh, Aditya and Goh, Gabriel and Agarwal, Sandhini and Sastry, Girish and Askell, Amanda and Mishkin, Pamela and Clark, Jack and others},
  booktitle={International conference on machine learning},
  pages={8748--8763},
  year={2021},
  organization={PmLR}
}

@article{tschannen2023image,
  title={Image captioners are scalable vision learners too},
  author={Tschannen, Michael and Kumar, Manoj and Steiner, Andreas and Zhai, Xiaohua and Houlsby, Neil and Beyer, Lucas},
  journal={Advances in Neural Information Processing Systems},
  volume={36},
  pages={46830--46855},
  year={2023}
}

@inproceedings{zhai2023sigmoid,
  title={Sigmoid loss for language image pre-training},
  author={Zhai, Xiaohua and Mustafa, Basil and Kolesnikov, Alexander and Beyer, Lucas},
  booktitle={Proceedings of the IEEE/CVF international conference on computer vision},
  pages={11975--11986},
  year={2023}
}

@article{wan2024locca,
  title={Locca: Visual pretraining with location-aware captioners},
  author={Wan, Bo and Tschannen, Michael and Xian, Yongqin and Pavetic, Filip and Alabdulmohsin, Ibrahim and Wang, Xiao and Pinto, Andr{\'e} S and Steiner, Andreas and Beyer, Lucas and Zhai, Xiaohua},
  journal={Advances in Neural Information Processing Systems},
  volume={37},
  pages={116355--116387},
  year={2024}
}

@inproceedings{xiao2024florence,
  title={Florence-2: Advancing a unified representation for a variety of vision tasks},
  author={Xiao, Bin and Wu, Haiping and Xu, Weijian and Dai, Xiyang and Hu, Houdong and Lu, Yumao and Zeng, Michael and Liu, Ce and Yuan, Lu},
  booktitle={Proceedings of the IEEE/CVF Conference on Computer Vision and Pattern Recognition},
  pages={4818--4829},
  year={2024}
}

@article{dutta2025vision,
  title={Vision-language models in diagnostic imaging: review of technical advances, clinical validation, and practical deployment},
  author={Dutta, Niharika and Bose, Kartik and Syailendra, Emir and Chu, Linda and Gupta, Pankaj},
  journal={International Journal of Medical Informatics},
  pages={106227},
  year={2025},
  publisher={Elsevier}
}

@article{rajpurkar2017chexnet,
  title={Chexnet: Radiologist-level pneumonia detection on chest x-rays with deep learning},
  author={Rajpurkar, Pranav and Irvin, Jeremy and Zhu, Kaylie and Yang, Brandon and Mehta, Hershel and Duan, Tony and Ding, Daisy and Bagul, Aarti and Langlotz, Curtis and Shpanskaya, Katie and others},
  journal={arXiv preprint arXiv:1711.05225},
  year={2017}
}

@article{akhter2023ai,
  title={AI-based radiodiagnosis using chest X-rays: A review},
  author={Akhter, Yasmeena and Singh, Richa and Vatsa, Mayank},
  journal={Frontiers in big data},
  volume={6},
  pages={1120989},
  year={2023},
  publisher={Frontiers Media SA}
}

@inproceedings{tong2024eyes,
  title={Eyes wide shut? Exploring the visual shortcomings of multimodal LLMs},
  author={Tong, Shengbang and Liu, Zhuang and Zhai, Yuexiang and Ma, Yi and LeCun, Yann and Xie, Saining},
  booktitle={Proceedings of the IEEE/CVF Conference on Computer Vision and Pattern Recognition},
  pages={9568--9578},
  year={2024}
}

@inproceedings{Hu_2023, series={KDD ’23},
   title={Expert Knowledge-Aware Image Difference Graph Representation Learning for Difference-Aware Medical Visual Question Answering},
   url={http://dx.doi.org/10.1145/3580305.3599819},
   DOI={10.1145/3580305.3599819},
   booktitle={Proceedings of the 29th ACM SIGKDD Conference on Knowledge Discovery and Data Mining},
   publisher={ACM},
   author={Hu, Xinyue and Gu, Lin and An, Qiyuan and Zhang, Mengliang and Liu, Liangchen and Kobayashi, Kazuma and Harada, Tatsuya and Summers, Ronald M. and Zhu, Yingying},
   year={2023},
   month=aug, pages={4156–4165},
   collection={KDD ’23} }

@inproceedings{huang2021gloria,
  title={Gloria: A multimodal global-local representation learning framework for label-efficient medical image recognition},
  author={Huang, Shih-Cheng and Shen, Liyue and Lungren, Matthew P and Yeung, Serena},
  booktitle={Proceedings of the IEEE/CVF international conference on computer vision},
  pages={3942--3951},
  year={2021}
}

@article{radford2019language,
  title={Language models are unsupervised multitask learners},
  author={Radford, Alec and Wu, Jeffrey and Child, Rewon and Luan, David and Amodei, Dario and Sutskever, Ilya and others},
  journal={OpenAI blog},
  volume={1},
  number={8},
  pages={9},
  year={2019}
}

@article{beyer2023study,
  title={A study of autoregressive decoders for multi-tasking in computer vision},
  author={Beyer, Lucas and Wan, Bo and Madan, Gagan and Pavetic, Filip and Steiner, Andreas and Kolesnikov, Alexander and Pinto, Andr{\'e} Susano and Bugliarello, Emanuele and Wang, Xiao and Yu, Qihang and others},
  journal={arXiv preprint arXiv:2303.17376},
  year={2023}
}

@article{wu2021chest,
  title={Chest imagenome dataset for clinical reasoning},
  author={Wu, Joy T and Agu, Nkechinyere N and Lourentzou, Ismini and Sharma, Arjun and Paguio, Joseph A and Yao, Jasper S and Dee, Edward C and Mitchell, William and Kashyap, Satyananda and Giovannini, Andrea and others},
  journal={arXiv preprint arXiv:2108.00316},
  year={2021}
}

\end{document}